# Handling uncertainty in a system for text-symbol context analysis.*

Bjørnar Tessem[†] & Lars Johan Ersland

Department of Informatics
University of Bergen
N-5007 Bergen, Norway

Email: bjornar@eik.ii.uib.no

## Abstract

In pattern analysis, information regarding an object can often be drawn from its surroundings. This paper presents a method for handling uncertainty when using context of symbols and texts for analyzing technical drawings. The method is based on Dempster-Shafer theory and possibility theory.

## 1 Introduction

In several pattern analysis problems it is interesting to look at the context in which an object is placed to gain information on what the object is. This article focuses on a method for analyzing one such problem, the problem of combining text and symbols in a system for processing technical drawings. In particular we look at the method for handling uncertainty in this system.

In our system we have a pattern recognition program that proposes alternatives for every text string and every symbol. To all strings or symbols the recognizer gives a probability distribution on the possible alternatives. Our problem is to combine these probabilities with information from the context of the strings and symbols.

This problem is essentially the same as combining uncertainty measures from different sources of evidence. One promising candidate for solving this kind of problems is Dempster-Shafer theory [5], and we shall use this theory to combine our probabilities with the uncertainty measure obtained from the context analysis. This uncertainty measure will be set equal to the plausibility measure of Dempster-Shafer theory. Results from [4] are used in the argumentation for the ideas behind our method.

In the next section a short review of Dempster-Shafer theory is given, and we show show how to combine the probabilities and plausibilities by using a result from [5]. In section 3 we discuss how the plausibilities are obtained and give an example of the use of this method.

---

*This paper is a revised version of the paper with the same title presented at the SCAI '88 conference in Tromso, Norway.

[†]This work was partly supported by the Royal Norwegian Council of Scientific and Industrial Research grant IT2.23.18229.

## 2  Combination of evidence using Dempster-Shafer theory.

The Dempster-Shafer theory of evidence was introduced in the seventies by Shafer [5] as an extension of probability theory and is based on Dempsters work from the sixties [2]. See [1,6,7] for a more extensive introduction to the theory.

Consider a set $\Theta$, called a *frame of discernment*, of possible values for a variable $q$. Evidence regarding the value of $q$ can be represented as a probability distribution on $2^\Theta$, the set of subsets of $\Theta$. Let us define the function $m$ that gives us this kind of probability distribution:

$$m : 2^\Theta \to [0,1]$$

such that

$$m(\emptyset) = 0 \text{ and } \sum_{A \subseteq \Theta} m(A) = 1.$$

This function is called a basic probability assignment (bpa). The value $m(A)$ may be interpreted as the evidence that $q$ takes a value in $A$. It is important to notice that the value $m(A)$ gives no evidence to the subsets of $A$ and also contributes nothing to the supersets of $A$. The sets $A$ such that $m(A) > 0$ are called the *focal elements* of $m$. Observe that when all the focal elements are singleton subsets of $\Theta$ we have a standard probability distribution on $\Theta$.

Now, let us define $Bel(A)$, a measure of total belief in $A$.

$$Bel(A) = \sum_{B \subseteq A} m(B).$$

By duality we have a plausibility function $Pl(A)$:

$$Pl(A) = \sum_{A \cap B \neq \emptyset} m(B).$$

We get

$$\begin{aligned} Pl(A) &= \sum_{A \cap B \neq \emptyset} m(B) = \\ &= 1 - \sum_{A \cap B = \emptyset} m(B) = \\ &= 1 - \sum_{B \subseteq A^C} m(B) = 1 - Bel(A^C). \end{aligned}$$

We may have evidence from two different sources regarding the value of a variable $q$ given in two bpas. The combination of two such bpas, $m_1$ and $m_2$ is done in the following way:

$$m_3(C) = m_1 \oplus m_2(C) = \begin{cases} \dfrac{\sum_{A \cap B = C} m_1(A) m_2(B)}{1 - \kappa} & C \neq \emptyset \\ 0 & C = \emptyset \end{cases}$$

where

$$\kappa = \sum_{A \cap B = \emptyset} m_1(A) \cdot m_2(B).$$

We see that this is a bpa since

$$\sum_A \sum_B m_1(A) m_2(B) - \sum_{A \cap B = \emptyset} m_1(A) m_2(B) =$$
$$\sum_A m_1(A) \sum_B m_2(B) - \kappa = 1 \cdot 1 - \kappa = 1 - \kappa.$$

Now, consider the problem of combining a symbol and a text string related to that symbol. We are not certain of the values of neither the symbol nor the text, but have a limited set of possibilities for each. From one source we have probabilities for the possible values of the symbol and probabilities for the possible values of the text. From another source we have a measure of how typical combinations of text and symbol are, the measure given as a number in the interval [0,1]. We want to combine the evidence from the two sources and get an uncertainty measure of the possible values for the combination of symbol and text.

We have a set $S = \{s_1, \ldots, s_n\}$ of different possible symbols, and a set $T = \{t_1, \ldots, t_m\}$ of different possible text strings. Our frame of discernment, $\Theta$ will thus be equal to the set $S \times T = \{(s,t) \mid s \in S, t \in T\}$. From the probabilities of symbols we construct a bpa in the following way:

We have from the probability distribution on $S$ that

$$m_1(\{(s,t) \mid t \in T\}) = P(s) \quad \text{for all } s \in S.$$

because $P(s)$ only gives evidence concerning the value of the symbol in the pair. Thus, $P(s)$ gives evidence regarding all pairs $(s,t), t \in T$. For all other subsets of $\Theta$, $m_1 = 0$. The same argument yields for the text strings:

$$m_2(\{(s,t) \mid s \in S\}) = P(t) \quad \text{for all } t \in T.$$

Let us now do the combination $m_3 = m_1 \oplus m_2$. The combination will involve only the focal elements of $m_1$ and $m_2$. From every such pair from $m_1$ and $m_2$ the intersection is the singleton $\{(s,t)\}$ and the combined value is $P(s)P(t)$. On all other subsets of $\Theta$, $m_3$ will take the value 0. Thus

$$m_3(A) = \begin{cases} P(s)P(t) & A = \{(s,t)\} \\ 0 & A \text{ is not a singleton.} \end{cases}$$

This is a standard probability distribution on the elements of $\Theta$ and gives us $P((s,t)) = P(s)P(t)$ which is the same result we would get from standard probability theory. Therefore, in this case Dempster-Shafer does not give us anything new.

Now look at the measure of typicallity we have from the other source of evidence. This is a function that assigns to every pair $(s,t)$ a value in [0,1]. If we now accept that this function can be interpreted as the plausibility function of the Dempster-Shafer framework we will show that the probability of a pair given the two sources of evidence, $P^*((s',t'))$ is

$$P^*((s',t')) = \frac{P((s',t'))Pl(\{(s',t')\})}{\sum_{s,t} P((s,t))Pl(\{(s,t)\})}$$

In order to accept the typicality function as plausibilities we must require that the sum of the typicality function of the elements must be greater than or equal to 1, since this is the case for the sum of the plausibilities of the singleton sets in a bpa. In the particular case where the plausibilities add up to 1 exactly, we have a standard probability distribution on $\Theta$.

Let us state the problem precisely: We have a frame of discernment $\Theta$ and two sources of evidence conserning the value of a variable $q$ in $\Theta$. From the first source we have a standard probability assignment $m_3$ on $\Theta$. From the second we have the plausibilities $Pl(A)$, for every singleton set $A$, corresponding to an unknown bpa $m_4$. We want to combine these two sources of evidence. The result shown below is from [5].

Let $m_5 = m_3 \oplus m_4$. Rewriting the definition of $\oplus$ slightly we get

$$m_5(A) = \frac{\sum_{B \cap C = A} m_3(B) m_4(C)}{\sum_{B \cap C \neq \emptyset} m_3(B) m_4(C)}$$

Here $m_3(B) = 0$ unless $B$ is a singleton. Hence $m_5(A)$ will be greater than 0 only if $A$ is a singleton and so $m_5$ correspond to a probability distibution $P^*$. We get

$$\begin{aligned} P^*(a) &= m_5(\{a\}) \\ &= \frac{\sum_{a \in C} m_3(\{a\}) m_4(C)}{\sum_{b \in C} m_3(\{b\}) m_4(C)} \\ &= \frac{m_3(\{a\}) \sum_{a \in C} m_4(C_j)}{\sum_b m_3(\{b\}) \sum_{b \in C} m_4(C)} \end{aligned}$$

By assumption $m_3(\{b\}) = P(b)$. Further

$$\sum_{b \in C} m_4(C) = \sum_{\{b\} \cap C \neq \emptyset} m_4(C) = Pl(\{b\}).$$

Hence

$$P^*(a) = \frac{P(a) Pl(\{a\})}{\sum_b P(b) Pl(\{b\})}.$$

In particular, for the symbol/text string problem we get

$$P^*((s', t')) = \frac{P((s', t')) Pl(\{(s', t')\})}{\sum_{s,t} P((s, t)) Pl(\{(s, t)\})}.$$

The generalisation to problems with many symbols and text strings is obvious. Our consern now is to find the plausibilities of a symbol-text string combination.

## 3  Aquirement of plausibilities.

In our system, the data we get from the context analyzer are fuzzy values on how typical a combination between one alternative symbol/text string and another symbol/text string is. That is, we have a number in $[0, 1]$ that tells us how possible a relation between a possible symbol/text string and another possible symbol/text string is. So what we have is a joint possibility distribution on $S \times T$. We want to use these possibilties as plausibilities of singleton sets of a bpa.

Why can we accept a possibility distribution as a plausibility function? Our answer lies in a relation between Zadeh's possibility distribution type evidence [8] and Dempster-Shafer type evidence. If we normalize Zadeh's possibility distribution by dividing all possibility values by the largest we get a possibility distribution with a maximum of 1. This is the type of possibility distribution Prade defines in [4].

Now let $\Pi(x)$ be the possibility of $x \in \Theta$, where $\Theta$ is our frame of discernment or in possibilistic terms, universe of discourse. Further let the possibility measure of a subset $B$ of $\Theta$ be $Poss(B) = max\{\Pi(x) : x \in B\}$. Prade shows that the plausibility $Pl(B)$ of a set $B \subseteq \Theta$ of a bpa where the focal elements $A_1, \ldots, A_n$ satisfy $A_1 \subset A_2 \subset \cdots \subset A_n = \Theta$ is equal to the possibility measure $Poss(B)$ of a possibility distribution where $\Pi(x) = Pl(\{x\})$. Reversely, it can be shown that for any normal possibility distribution there is exactly one bpa that satisfies $Poss(B) = Pl(B)$ for any $B \subseteq \Theta$. This bpa will have the same nested structure as above and is easily constructed from the possibility distribution. This shows that these two ways of representing uncertainty is equivalent.

When seeing the relation between the representation of uncertainty in a possibility distribution and a nested structure bpa we don't see any problems in using the normalized possibility values as plausibilities of the singleton sets of a bpa. The normalization done in the equation for combining probabilities and plausibilities does in fact make any prior normalization unnecessary, so what we really use in the combination of probabilities and plausibilities is the fuzzy values obtained for the typicallity of a possible connection.

We of course use the same argument when considering the general case of many symbols and strings. However, we need a method for obtaining fuzzy values in the general case. To solve this problem we use the joint possibility distrubution on symbols and text and combine these to get a possibility distribution on a larger structure.

The context analyzer finds possible connections between symbols and text strings by using rules like:

*If a symbol is the symbol for resistance and a text string ends in 'k$\Omega$' then there is possibility 1 that they are connected.*

This kind of rules together with distance between symbol and text, relative position and other features in the original drawing then gives us the fuzzy value of this possible connection.

We introduce a graph, which we call the complete connection graph, that contains all symbols $s_i$ and text strings $t_j$ as vertices and possible connections between them as edges. On every edge we have a value $v(e)$ that gives us the possibility measure on that edge. If the value of an edge is 0 this edge is not included in the complete connection graph.

However, typically not all the connections in the complete connection graph do really exist. Therefore we want to find a subgraph of this complete graph that contains the "best" possible connections. We call this subgraph a solution graph of the complete connection graph. To simplify the problem of finding a solution graph we have put some constraints on the graph. It is, for instance, very rare that a text string is connected to more than one symbol, and it is also rarely connected other text strings. Therefore we say that in a solution graph a text string node must be of degree 1. Symbols, however, may be connected to any number of strings and symbols. In particular a symbol may be isolated. When this is the case, an edge from the symbol to itself is added to the graph. Its value is equal to how typical it is for the symbol to be isolated. Thus, any s-vertex is of degree at least 1 in the solution graph.

The value of this solution graph will we define to be the minimum value over all the edges in the graph. And it is this value we will use as a plausibility when combining with probabilities. In figure 1 we show a complete connection graph and its solution graph.

Let us give a formal definition of a solution graph. We define $v(G)$, the value of a subgraph $G$, by

$$v(G) = \min\{v(e) \mid e \text{ is an edge in } G\}.$$

Vertices that represent text strings are called t-vertices, and vertices that represent symbols are called s-vertices. A *solution graph* $G$ is a subgraph having the same vertex set as the complete

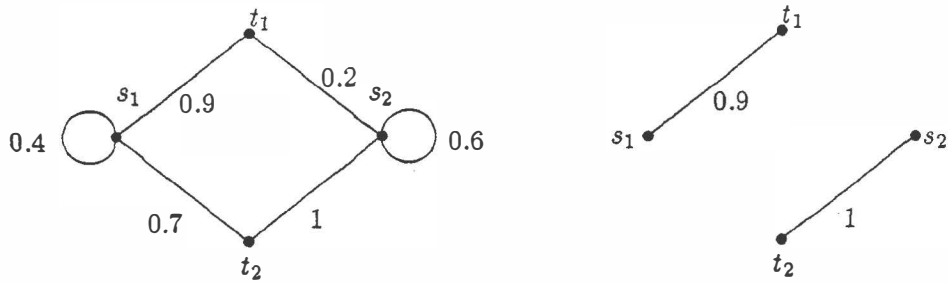

Figure 1: A complete connection graph and its solution graph.

connection graph and such that

1. All t-vertices are of degree 1.

2. An s-vertex has either an edge to itself as its only edge connected to it, or one or more edges to other vertices, excluding itself.

3. $v(G)$ is maximal.

An algorithm for finding the value is shown in figure 2, and a proof for the correctness of the algorithm follows:

The complete connection graph given as input to the algorithm, may not contain a solution graph. In that case the algorithm returns 0 in step 1, because this combination of values is not possible by the constraints we have put on the degrees of the vertices. Let $P = \{$The remaining graph contains a solution graph$\}$. We want to show that $P$ is true each time the algorithm enters the outer loop. $P$ is clearly true the first time the outer loop is entered. Now suppose $P$ is true before entering step 3 at any time during the algorithm.

We consider two possibilities when we enter step 3. The first possibility is that the edge $e$ with the smallest graph is the single edge connected to one of its endpoints. This edge then has to be contained in any solution graph. Hence $v(G) = v(e)$ for a solution graph $G$. We return from the algorithm with this value in step 4. The second possibility is that both endpoints of $e$ has degree at least 2. Then $e$ is not a part of any solution graph. In this case we continue to step 5. We have to show that $P$ is invariant for the inner loop starting at step 5.

In step 6 we look at an unset vertex $n$ of degree 1 in the remaining graph. The single edge from $n$ in the remaining graph is the edge from $n$ to $m$, and this must be contained in any solution graph. By condition 2, the edge $(m, m)$ can not be part of any solution graph. If $m$ is a t-vertex, $(m, n)$ is the only edge from $m$ in a solution graph, by condition 1. Hence, edges removed in step 6 are not part of any solution graph. Therefore $P$ is true at the exit of the inner loop.

This proves that $P$ is true when the outer loop completes an iteration and starts the next. To see that the algorithm terminates, observe that at least one edge is removed in each iteration of the outer loop, and that in the inner loop at least one vertex is marked 'set' in each iteration. This concludes our proof of the algorithm.

The algorithm solves the problem of finding a plausibility in time of order equal to the number of edges in the complete connection graph. If the number of symbols and text strings considered are $n$, then the number of edges is at most of order $n^2$. In our problem, however, the number can be considered to be linear because of the planar nature of graphs constructed from technical drawings.

The important question conserning time is the number of complete connection graphs one has to compute plausibilities for. It is easy to see that this number is exponential in the number of

> **algorithm FIND-COMBINED-POSSIBILITY.**
>
> 1. Construct the graph from the typicallity values, using an adjacency list representation. Mark all vertices to 'unset'. If some vertex has degree 0, return 0.
>
> 2. Loop.
>
> 3. Remove the edge $e$ with smallest value $v(e)$ from the graph.
>
> 4. If some adjacency list is empty, return the value of the last edge deleted.
>
> 5. Loop-while some 'unset' vertex $n$ has an adjacency list with only one element do step 6.
>
> 6. 
>    - If $n$ is a t-vertex, the single vertex in its adjacency list is an s-vertex $m$. Remove $m$ from $m$'s adjacency list. Mark $n$ 'set'.
>    - If $n$ is an s-vertex, there are three possibilities for the single vertex $m$ in $n$'s adjacency list. If $m = n$ do nothing, if $m$ is another s-vertex, delete $m$ from $m$'s adjacency list, if $m$ is a t-vertex delete all edges from $m$ except the one to $n$. In all cases mark $n$ 'set'.
>
> 7. End loop-while.
>
> 8. End loop.

Figure 2: An algorithm for finding the value of a solution graph.

symbols and text strings in the drawing. The way this problem is solved in our system is to consider at a time only a subgraph of symbols and text strings that are, for no value of symbols and text strings connected to vertices outside the subgraph. In this case, the work reduces to computing plausibilities for each subgraph and afterwards combine the results from every subgraph. This is of course not possible in all cases, and other simplifications should be considered. An approximation would be to delete all edges with values lower than a certain limit. One could then use the methods mentioned above. More work has to be done to observe consequences of this approximation.

## 4 Conclusion.

We have presented a method for combining probabilities from obtained from character and symbol representation with results from context analysis in a pattern recognition system. The numerical methods used in the system are based on ideas from both possibility theory and Dempster-Shafer theory and shows that it is possible to combine these in practical applications. The results, however, depends on our willingness to accept possibilities as Dempster-Shafer plausibilities. In our opinion this is a reasonable assumption given the close relationship shown to exist between possibility distribution and nested structure bpas. Our largest problem is the combinatorial explosion that occurs when adding new symbols or text strings to the drawing. Further work has to be done on that problem.